\documentclass{article}

\usepackage[preprint]{neurips_2024}

\usepackage{subfig}
\usepackage{graphicx}
\usepackage{amsmath}
\usepackage[utf8]{inputenc} 
\usepackage[T1]{fontenc}    
\usepackage{hyperref}       
\usepackage{url}            
\usepackage{booktabs}       
\usepackage{amsfonts}       
\usepackage{nicefrac}       
\usepackage{microtype}      
\usepackage{xcolor}         
\usepackage{courier}

\title{Revisiting MoE and Dense Speed-Accuracy Comparisons for LLM Training}

\author{
Xianzhi Du \quad Tom Gunter \quad Xiang Kong \quad Mark Lee \quad \textbf{Zirui Wang} \\
 \quad \textbf{Aonan Zhang} \quad \textbf{Nan Du} \quad \textbf{Ruoming Pang} \\
Apple\\
\texttt{\{xianzhi,r\_pang\}@apple.com}\\
}

\begin{document}
\maketitle

\begin{abstract}
Mixture-of-Experts (MoE) enjoys performance gain by increasing model capacity while keeping computation cost constant.
When comparing MoE to dense models, prior work typically adopt the following setting: 1) use FLOPs or activated parameters as a measure of model complexity; 2) train all models to the same number of tokens. We argue that this setting favors MoE as FLOPs and activated parameters do not accurately measure the communication overhead in sparse layers, leading to a larger actual training budget for MoE.
In this work, we revisit the settings by adopting step time as a more accurate measure of model complexity, and by determining the total compute budget under the Chinchilla compute-optimal settings.
To efficiently run MoE on modern accelerators, we adopt a 3D sharding method that keeps the dense-to-MoE step time increase within a healthy range.
We evaluate MoE and dense LLMs on a set of nine 0-shot and two 1-shot English tasks, as well as MMLU 5-shot and GSM8K 8-shot across three model scales at 6.4B, 12.6B, and 29.6B. Experimental results show that even under these settings, MoE consistently outperform dense LLMs on the speed-accuracy trade-off curve with meaningful gaps. Our full model implementation and sharding strategy has been released at~\url{https://github.com/apple/axlearn}.

\end{abstract}

\section{Introduction}
\label{intro}
Recently, MoE has shown promising results on language (\cite{zoph2022stmoe,pmlr-v162-du22c,NEURIPS2022_erc,fedus2022switch,jiang2024mixtral,komatsuzaki2023sparseupcycle,shen2023flanmoe,dai2024deepseekmoe}), multimodal (\cite{mustafa2022multimodal,lin2024moellava}) and computer vision (\cite{ruiz2021scaling,komatsuzaki2023sparseupcycle,daxberger2023mobile,Chen_2023_ICCV}) tasks. By decoupling computation cost from model scale, MoE scales model capacity without affecting computation cost. 

When comparing MoE to dense models, it is common in existing work to use FLOPs or activated model parameters as a measure of a model's computation cost (\cite{ruiz2021scaling,jiang2024mixtral,pmlr-v162-du22c,shen2023flanmoe,NEURIPS2022_erc,dai2024deepseekmoe,lin2024moellava}). However, as MoE sparsity grows, the communication overhead during routing (e.g. the \texttt{all2all} and \texttt{allreduce} communication primitives) also increases. Such communication overhead cannot be accurately captured by FLOPs or number of activated parameters, leading to a setting that favors MoE. Furthermore, existing work commonly adopt an identical training recipe for MoE and dense models, i.e. utilizing the same batch size and same number of training steps, resulting in a higher total effective computation cost for MoE.

In this work, we propose an alternative setting for comparing MoE and dense models under the modern LLM training paradigm. Specifically, we propose two main revisions:
1) Use step time to measure model complexity which fully captures the communication overhead in MoE layers. The step time for MoE and dense models are measured and optimized under identical settings.
2) Adopt the Chinchilla (\cite{hoffmann2022training}) compute-optimal setting of a 20:1 token-to-parameter ratio, a setting that is optimized for dense LLM training, to decide the total training budget for MoE vs. dense comparison at multiple model scales.

To optimize the MoE training step time, we adopt the GShard (\cite{lepikhin2021gshard}) method and build our model on top of GSPMD (\cite{xu2021gspmd}). We further adopt a 3D sharding strategy that partitions a model along \texttt{Data}, \texttt{Expert}, and \texttt{Model} axes. The final MoE implementation efficiently runs on modern accelerators and is capable of utilizing more experts without a significant degradation in step time. Across a wide range of scales, the 3D sharding strategy keeps the dense-to-MoE step time increase within 20\%.

We compare MoE and dense LLMs across a wide range of scales at 6.4B, 12.6B, 29.6B and evaluate the models on a rich set of LLM benchmarks including nine 0-shot and two 1-shot English tasks covering common sense reasoning, question answering and reading comprehension, as well as MMLU 5-shot and GSM8K 8-shot. Extensive evaluations show that under these challenging settings, MoEs consistently outperform dense LLMs on the speed-accuracy trade-off curve.



\section{Methods}
\subsection{Architecture}
The dense LLM architecture used in this paper generally follows LLaMA2 (\cite{touvron2023llama2}). Specifically, we adopt the following transformer architectural modifications: 1) Pre-normalization and RMSNorm (\cite{zhang-rmsnorm}); 2) SwiGLU (\cite{shazeer2020glu}) as the activation function; 3) RoPE (\cite{su2023rope}) as the positional embedding. Grouped-query attention (\cite{ainslie2023gqa}) is not used as our experiments focus on training.

Our MoE architecture shares the same architecture as the dense LLM, except replacing FFNs with their sparse counterparts in MoE layers. We also make the following design decisions:

\textbf{Number of sparse layers.} More sparse layers leads to higher model sparsity and generally better performance. On the other hand, it increases the model's total parameters and computation cost. Typical choices for this hyper-parameter include Every-$K$ (\cite{jiang2024mixtral,zoph2022stmoe,pmlr-v162-du22c}) or Last-$K$ (\cite{ruiz2021scaling,komatsuzaki2023sparseupcycle}). We adopt the Every-4 setting in our experiments as it provides a better speed-accuracy trade-off. Specifically, we replace the last dense layer out of every 4 layers with a sparse layer.

\textbf{Number of experts.} More experts leads to a higher model capacity while keeping the model's activated parameters constant. Existing work (\cite{pmlr-v162-du22c,fedus2022switch,clark2022unified}) shows that using more experts leads to monotonic performance improvement, which gradually diminishes when the number grows beyond $256$. In this paper, we experiment with different numbers of experts and show that with our 3D sharding strategy, more experts leads to a better performance without affecting step time. 

\textbf{Routing method and expert capacity.} We adopt the Top-$K$ routing (\cite{shazeer2017moe}) for autoregressive modeling with $K=2$ and expert capacity $C=2$ in all our experiments. Other common routing methods include Top-1 (\cite{fedus2022switch}) and expert-choice routing (\cite{NEURIPS2022_erc}).

\textbf{Auxiliary loss.} To encourage better expert load balancing for Top-$K$ routing, we adopt a load balancing loss with a coefficient of $0.01$ (\cite{lepikhin2021gshard,pmlr-v162-du22c}). We also adopt the router z-loss proposed in ST-MoE (\cite{zoph2022stmoe}) with a coefficient of $0.001$, which we found helps with stabilizing large MoE training.

\begin{table}
  \caption{An overview of our MoE transformer sharding specifications for dense and sparse layers. \texttt{None} means no sharding is performed on the dimension.}
  \vspace{2mm}
  \label{tab:sharding}
  \centering
  \begin{tabular}{l c | c }
    \toprule
    name & type & sharding specification \\
    \midrule
    attention & dense weights & (\texttt{Expert}, \texttt{Model}) \\
    $\text{FFN}_1$ & dense weights & (\texttt{Expert}, \texttt{Model}) \\
    $\text{FFN}_2$ & dense weights & (\texttt{Model}, \texttt{Expert}) \\
    $\text{FFN}_1$ & dense activation & ((\texttt{Data}, \texttt{Expert}), \texttt{None}, \texttt{Model}) \\
    $\text{FFN}_2$ & dense activation & ((\texttt{Data}, \texttt{Expert}), \texttt{None}, \texttt{Model}) \\
    \midrule
    router, ME & MoE weights & (\texttt{None}, \texttt{None})  \\
    $\text{FFN}_1$, EMH & MoE weights &  (\texttt{Expert}, \texttt{None}, \texttt{Model}) \\
    $\text{FFN}_2$ EHM & MoE weights & (\texttt{Expert}, \texttt{Model}, \texttt{None}) \\
    OGSM & MoE activation & (\texttt{Data}, \texttt{Expert}, \texttt{None}, \texttt{Model}) \\ 
    OGSEC & MoE activation &  (\texttt{Data}, \texttt{Expert}, \texttt{None}, \texttt{None}, \texttt{None}) \\
    OEGCM & MoE activation & (\texttt{Data}, \texttt{Expert}, \texttt{None}, \texttt{None}, \texttt{Model}) \\
    OGECM & MoE activation & (\texttt{Data}, \texttt{Expert}, \texttt{None}, \texttt{None}, \texttt{Model}) \\
    OEGCH & MoE activation & (\texttt{Data}, \texttt{Expert}, \texttt{None}, \texttt{None}, \texttt{Model}) \\
    \bottomrule
  \end{tabular}
\end{table}

\subsection{3D sharding and MoE Layer Implementation}
As we use step time as a measurement for model complexity, optimizing the sparse MoE sharding strategy lies at the very foundation. In this work, we adopt a 3D sharding strategy across 3 axes: (\texttt{Data}, \texttt{Expert}, \texttt{Model}).

\texttt{\textbf{Data.}} The \texttt{Data} axis is used for data parallelism only. Along this axis, data is evenly partitioned and model weights are fully replicated. This axis is commonly used for inter-slice parallelism, where communication often happens over slower data center networking (DCN).

\texttt{\textbf{Expert.}} The \texttt{Expert} axis is designed for sharding the experts in sparse FFNs. To optimize the compute-to-memory ratio, we place at most one expert per core. The \texttt{Expert} axis provides flexibility and efficiency when scaling the number of experts in one model while keeping step time constant. Note that for dense layers, the \texttt{Expert} axis is used as the mesh axis for fully-sharded data parallelism (\cite{zhao2023fsdp}) (FSDP) which shards model parameters across data-parallel cores.

\texttt{\textbf{Model.}} The \texttt{Model} axis is used to shard attention heads and FFN hidden dimensions. This axis incurs heaviest communication among all of the axes and is typically used for intra-slice model parallelism, where communication happens over fast inter-core interconnect (ICI).

GShard (\cite{lepikhin2021gshard}) provides an efficient MoE layer implementation that extensively uses Einsum notations to express gating, dispatching and combining operations. In this work, we adopt the GShard implementation provided in the Praxis library\footnote{https://github.com/google/praxis}, as well as the outer batch trick used in Mesh-Tensorflow\footnote{https://github.com/tensorflow/mesh/blob/master/mesh\_tensorflow/transformer/moe.py}. Specifically, the input tokens to the MoE layer are evenly divided into $O$ outer batches, resulting in an additional leading dimension of shape $O$ in all activations. We then shard the outer batch dimension along the \texttt{Data} axis for best efficiency. Table~\ref{tab:sharding} provides an overview of our MoE transformer sharding specifications.

\subsection{Training Compute Budget and Model Design}
We evaluate MoE and dense LLMs speed-accuracy comparisons across 3 scales at 6.4B, 12.6B and 29.6B. At each scale, we adopt the Chinchilla token-to-parameter ratio (\cite{hoffmann2022training}) of $20$ : $1$ to determine the number of training tokens for the corresponding dense model. The total compute budget is determined by multiplying the training step time of the dense model and the total number of training steps. Given this budget, we design MoEs and determine the step time and the training steps, fixing the batch size and hardware accelerators. Unlike previous work that constructs MoE from the same scale dense backbone~\footnote{MoE's dense backbone is the dense model from which the MoE is constructed. For example, 1.2B dense is the backbone of 1.2B/64E.}, we trade model size for training tokens and design MoEs from smaller backbones. Ablations in Sec.~\ref{sec:wider_design} shows that MoEs designed with this method gives a better speed-accuracy trade-off curve. Table~\ref{tab:train_budget} provides the final train budget for all models used in our experiments.

\begin{table}
  \caption{Dense and MoE LLM comparisons at 6.4B, 12.6B and 29.6B scales. Note that the step time is only comparable for models within each scale due to being measured on different devices. Note that the 6.4B scale comparison uses CoreEN (0S) while other scales use the CoreEN (all).}
  \vspace{2mm}
  \label{tab:main_results}
  \centering
  \begin{tabular}{l | c | l | c  c  c}
    \toprule
    model & $\text{params}_{\text{act}}$ & step time & CoreEN & MMLU & GSM8K \\
    \midrule
    6.4B & 6.4B & 1.69s & 63.25 (0S) & - & - \\
    1.6B/256E & 2.1B & \textbf{0.82s} (\textcolor{red}{-51.5\%}) & 65.00 (0S) & - & - \\
    4.8B/256E & 6.2B & 1.41s (\textcolor{red}{-16.7\%}) & \textbf{65.40} (0S) & - & - \\
    \midrule
    12.6B & 12.6B & 2.94s & 57.25 (all) & 24.84 & 5.08 \\
    4.5B/256E & 5.3B & \textbf{1.50s} (\textcolor{red}{-49.0\%}) & 60.62 (all) & 29.52 & \textbf{12.13} \\
    8.1B/256E & 9.4B & 2.60s (\textcolor{red}{-11.2\%}) & \textbf{61.38} (all) & \textbf{30.18} & 11.97 \\
    \midrule
    29.6B & 29.6B & 6.56s & 62.73 (all) & 47.03 & \textbf{18.88} \\
    6.4B/64E & 7.5B & \textbf{1.85s} (\textcolor{red}{-71.8\%}) & \textbf{63.20} (all) & \textbf{48.37} & 16.91 \\
    \bottomrule
  \end{tabular}
\end{table}

\section{Experiments}

\subsection{Experimental Setup}
We use a training corpus with a similar data mixture as LLaMA2 (\cite{touvron2023llama2}). All models are trained with the AdamW optimizer (\cite{loshchilov2019decoupled}) with $\beta_1=0.9$, $\beta_2=0.95$, and $\epsilon = 10^{-5}$. We adopt a cosine learning rate schedule with 2000 warmup steps and decay to 10\% of the peak learning rate. We use a weight decay of $0.1$ and gradient clipping of $1.0$. We use the SentencePiece tokenizer (\cite{kudo2018sentencepiece}) with the Byte-Pair Encoding algorithm (\cite{sennrich2016bpe}). We train our small scale models on TPU-v5e devices and large scale models on TPU-v4 devices.

The MoE and dense LLMs are evaluated on a wide range of benchmarks:

\textbf{Core English tasks (CoreEN).} We evaluate the performance on common sense reasoning, reading comprehension and question answering benchmarks including ARC Easy and Challenge (0-shot) (\cite{clark2018arc-e-c}), HellaSwag (0-shot) (\cite{zellers2019hellaswag}), WinoGrande (0-shot) (\cite{sakaguchi2019winogrande}), PIQA (0-shot) (\cite{bisk2019piqa}), SciQ (0-shot) (\cite{sap2019socialiqa}), LAMBADA (0-shot) (\cite{paperno2016lambada}), TriviaQA (1-shot) (\cite{joshi2017triviaqa}) and WebQS (1-shot) (\cite{Berant2013webqs}). In our experiments, we report both the average performance on the seven 0-shot tasks (denoted as CoreEN (0S)) and the average performance on the whole nine tasks (denoted as CoreEN (all)). 

\textbf{MMLU.} We report the 5-shot performance on MMLU (\cite{mmlu}).

\textbf{Mathematical reasoning.} We report the 8-shot performance on GSM8K (\cite{cobbe2021gsm8k}). 

\subsection{Pretraining Results}
Table~\ref{tab:main_results} presents the pretraining results of dense and MoE LLMs at 6.4B, 12.6B and 29.6B scales on CoreEN, MMLU, and GSM8K. We can see that at any given scale, one or more MoEs outperform the dense baseline on the three benchmarks while being faster in training step time. 

\textbf{6.4B scale comparison.} The 6.4B dense baseline achieves $63.25\%$ accuracy on CoreEN (0S) average and runs at $1.69$ second per step on 512 TPU-v4 cores. The two MoE models we tested, 1.6B/256E and 4.8B/256E, both outperform the 6.4B dense while being faster in step time. In particular, the 1.6B/256E model achieves $+1.75\%$ on CoreEN (0S) while being $2.06\times$ as fast as the dense 6.4B. Due to insufficient model capacity and train tokens at this scale, all the models fail to show meaningful results on MMLU 5-shot or GSM8K 8-shot.

\textbf{12.6B scale comparison.} The 12.6B dense baseline achieves $57.25\%$ on CoreEN (all) and runs at $2.94$ second per step on 1024 TPU-v4 cores. It fails to show meaningful results on MMLU or GSM8K. The two MoEs we evaluated, 4.5B/256E and 8.1B/256E, both achieve better results on CoreEN (all) while being much faster in step time. At this scale, we further observe that the MoEs begin to achieve meaningful results on MMLU and GSM8K. In particular, the 4.5B/256E model achieves $+3.37\%$ on CoreEN average, $+4.68\%$ on MMLU and $+7.05\%$ on GSM8K while being $1.96\times$ as fast as the dense 12.6B.

\textbf{29.6B scale comparison.} The 29.6B dense baseline achieves $62.73\%$ on CoreEN (all), $47.03\%$ on MMLU and $18.88\%$ on GSM8K. The 6.4B/64E MoE attains very close performance on the three benchmarks: $+0.47\%$ on CoreEN (all), $+1.34\%$ on MMLU, and $-1.97\%$ on GSM8K, while being $3.55\times$ as fast as the baseline.

\begin{figure}%
    \centering
    \subfloat{{\includegraphics[width=6.5cm]{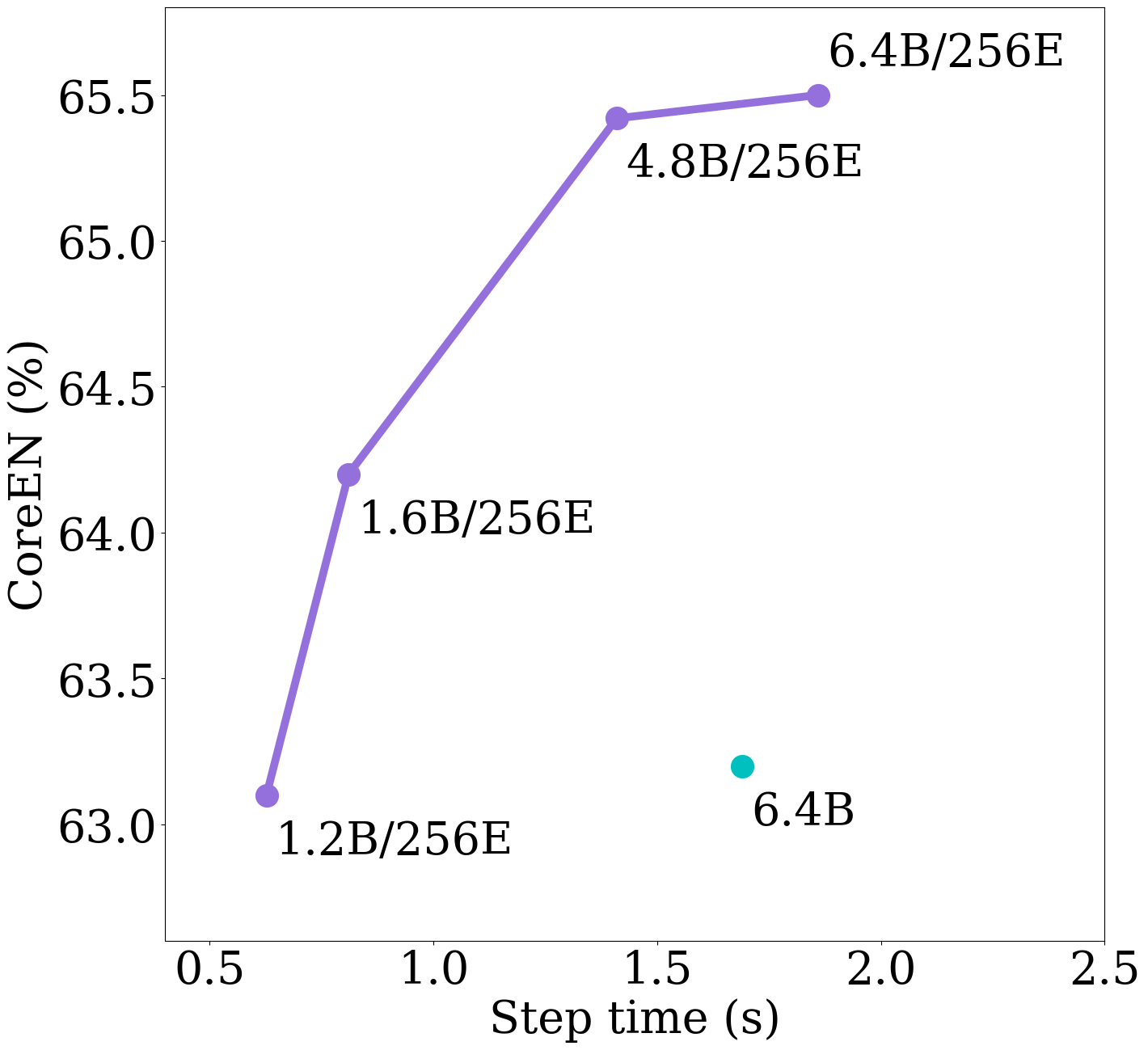} }}%
    \qquad
    \subfloat{{\includegraphics[width=6.5cm]{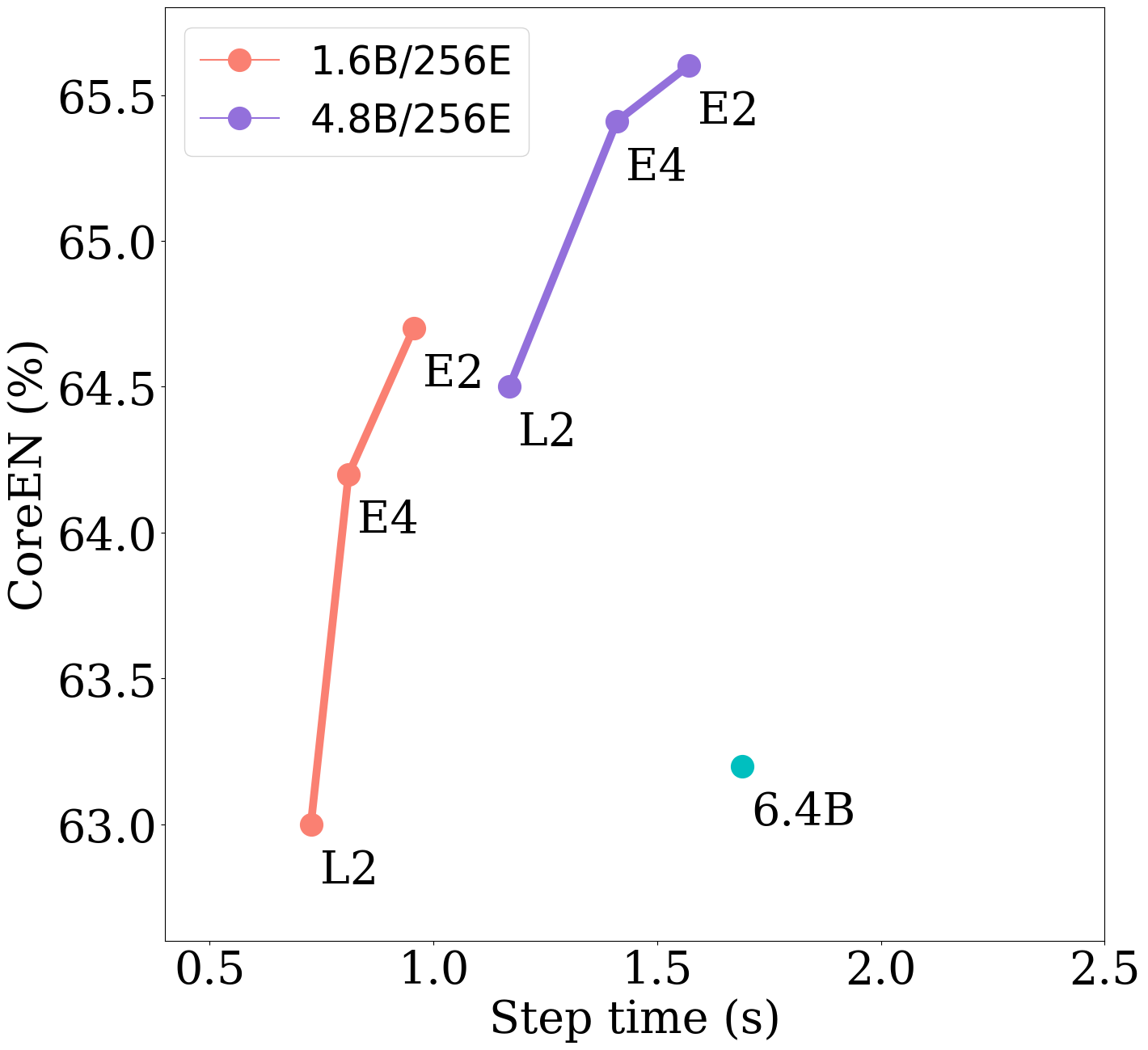} }}%
    \caption{Comparing MoEs designed from a wider range of choices to the 6.4B dense. We vary the dense backbone scale and the number of sparse layers. On the right figure, L2, E4, E2 represent last-2, every-4, every-2 number of layers, respectively.}%
    \label{fig:moe_arc}%
\end{figure}

\subsection{Explore a Wider Range of MoE Architecture Designs}
\label{sec:wider_design}
In this section, we show that MoE architectures constructed from a wide range of design choices outperform the dense baseline. Aside from the architectures we used in the main results, we design MoEs by using a wider range of dense backbone scales and different number of MoE layers. We fix the train budget for all ablations to the 6.4B dense Chinchilla budget and evaluate all models on the CoreEN (0S) benchmark using 512 TPU-v4 cores.

\textbf{Vary dense backbone scale.} Under this setting, we fix the number of MoE layers to every-4 and number of experts to 256 then vary the dense backbone scale with 1.2B, 1.6B, 4.8B and 6.4B. Figure~\ref{fig:moe_arc} (left) shows that the MoEs form a speed-accuracy trade-off curve that outperform the 6.4B dense with a meaningful gap. The trade-off curve also suggests that more MoE architectures interpolated onto this curve should outperform the 6.4B dense. The ablations also suggest that designing MoEs using smaller scale backbones gives better results than using the same dense scale backbone. In particular, the 4.8B/256E MoE underperforms the 6.4B/256E MoE by only 0.08\% on CoreEN (0S) while reducing the step time by 24.2\%. One hypothesis is that the equivalent dense scale of a MoE lies at somewhere between its activated parameters and its total parameters. Under the current dense Chinchilla setting, designing MoEs from the same scale dense backbone surely makes its equivalent dense scale being away from the Chinchilla compute-optimal frontier. We will leave the full study on the compute-optimal frontier for MoE as a future work.

\textbf{Vary number of MoE layers.} Under this setting, we use 1.6B and 4.8B as the dense backbones and vary the number of MoE layers with last-2, every-4 and every-2. Figure~\ref{fig:moe_arc} (right) shows that MoEs at the 1.6B and 4.8B backbone scales form speed-accuracy trade-off curves that outpform the 6.4B dense with a meaningful gap. The curves further suggest that MoEs constructed from more design choices interpolated onto the curves should outperform the dense baseline.

\begin{table}
  \caption{The effectiveness of scaling number of experts on the 1.6B MoEs. All models are trained with the 6.4B dense Chinchilla budget on 256 TPU-v5e cores. Mesh shape is ordered as (\texttt{Data}, \texttt{Expert}, \texttt{Model}).}
  \vspace{2mm}
  \label{tab:num_experts1}
  \centering
    \begin{tabular}{l | c  c | c}
    \toprule
    model & mesh shape & step time & CoreEN (0S) \\
    \midrule
    1.6B/16E & (16, 16, 1) & 1.07s & 62.48 \\
    1.6B/64E & (4, 64, 1) & 1.05s & 63.68 \\
    1.6B/256E & (1, 256, 1) &  1.07s & 65.0 \\
    \bottomrule
   \end{tabular}
\end{table}

 \begin{table}
  \caption{The effectiveness of scaling number of experts on the 4.5B and 8.1B MoEs. All models are trained with the 12.6B dense Chinchilla budget on 1024 TPU-v4 cores. Mesh shape is ordered as (\texttt{Data}, \texttt{Expert}, \texttt{Model}).}
  \vspace{2mm}
  \label{tab:num_experts2}
  \centering   
   \begin{tabular}{l | c c | c c c}
    \toprule
    model & mesh shape & step time & CoreEN (all) & MMLU & GSM8K \\
    \midrule
     4.5B/64E & (8, 64, 1) &  1.50s & 58.60 & 26.68 & 6.75 \\
    4.5B/256E & (2, 256, 1) & 1.50s & 60.62 & 29.52 & 12.13 \\
    \midrule
    8.1B/64E & (8, 64, 1) & 2.57s & 59.60 & 28.39 & 8.64 \\
    8.1B/256E & (2, 256, 1) & 2.60s & 61.38 & 30.18 & 11.97 \\
    \bottomrule 
  \end{tabular}
\end{table}

\subsection{Effectiveness of the 3D sharding method}\label{ablation:4d_sharding}
Our 3D sharding method brings two major benefits. First, it allows us to scale the number of experts without affecting train step time much. Second, it controls the communication overhead from routing in MoE layers within a healthy range.

\textbf{Scale number of experts without affecting step time.} Table~\ref{tab:num_experts1} and Table~\ref{tab:num_experts2} show the effectiveness of scaling number of experts with the 3D sharding method. We evaluate the 1.6B MoEs using the 6.4B scale train budget on 256 TPU-v5e cores and the 4.5B and 8.1B MoEs using the 12.6B scale budget on 1024 TPU-v4 cores. We see that scaling number of experts monotonically improves model performance on all three CoreEN, MMLU and GSM8K benchmarks while not affecting train step time much. To optimize model performance for training, the results suggest it would be beneficial to scale the number of experts to the limit of device efficiency\footnote{We consider two limits: 1) up to one expert per core; 2) device memory.} and model architecture\footnote{Existing work (\cite{fedus2022switch,pmlr-v162-du22c,clark2022unified}) shows that scaling number of experts beyond 256 gives diminished return.}.

\textbf{Controlled MoE communication overhead.} We evaluate the dense-to-MoE step time increase across a wide range of model scales with 85M, 1.2B and 6.4B. Table~\ref{tab:overhead} summaries the results. We can see that the 3D sharding method controls the dense-to-MoE step time increase within a healthy range, typically below 20\%. It is worth noting that for one extreme setup where the dense model just fits into the device memory without sharding along the \texttt{Model} axis but MoEs would run out of memory and have to shard along the \texttt{Model} axis. In this case the step time increase would become much larger and we suggest either reducing batch size or using more devices to control the step time increase.

\textbf{Our sharding method vs. other solutions.} We compare our sharding method to two other solutions under the GShard MoE implementation. 1) The conventional 2D sharding method. When the number of experts is smaller than the number of devices, we have to shard the expert dimension of the MoE weights and activations along both \texttt{Data} and \texttt{Model} axes, leading to suboptimal efficiency; 2) Another 2D method is to pad the expert dimension to the number of devices so it can be sharded only along \texttt{Data}. This is equivalent to adding more experts that would never process any tokens. We evaluate the three methods using a 1.6B/64E model trained with the 6.4B budget on 256 TPU-v5e cores. Table~\ref{tab:outer_batch} shows the step time comparisons. 

\begin{table}
  \caption{Step time increase from dense to MoEs across a wide range of model scales. Mesh shape is ordered as (\texttt{Data}, \texttt{Expert}, \texttt{Model}).}
  \vspace{2mm}
  \label{tab:overhead}
  \centering
  \begin{tabular}{l | c  c  c | c c}
    \toprule
    model & devices & batch size & mesh shape & step time & delta \\
    \midrule
    85M & 64 TPU-v5e & 25k & (64, 1, 1) & 0.50s & - \\
    85M/32E & 64 TPU-v5e & 25k & (2, 32, 1) & 0.53s & +6.0\% \\
    85M/64E & 64 TPU-v5e & 25k & (1, 64, 1) & 0.54s & +8.0\% \\
    \midrule
    1.2B & 256 TPU-v5e & 1M & (256, 1, 1) & 0.69s & - \\
    1.2B/256E & 256 TPU-v5e & 1M & (1, 256, 1) & 0.81s & +17.4\% \\
    \midrule
    6.4B & 512 TPU-v4 & 1M & (1, 256, 1) & 1.69s & - \\
    6.4B/64E & 512 TPU-v4 & 1M & (4, 64, 1) & 1.92s & +13.6\% \\
    6.4B/256E & 512 TPU-v4 & 1M & (1, 256, 1) & 1.86s & +10.1\% \\
    \bottomrule
  \end{tabular}
\end{table}

\begin{table}
  \caption{Effectiveness of our final sharding specifications compared to naive 2D sharding or 2D sharding with padding. All numbers are measured on 256 TPU-v5e cores using a 1.6B/64E MoE.}
  \vspace{2mm}
  \label{tab:outer_batch}
  \centering
  \begin{tabular}{l | c | c  }
    \toprule
     & $\text{params}_{\text{total}}$ & step time  \\
    \midrule
    2D sharding (naive) & 23.8B & 1.98s \\
    2D sharding (padding) & 46.6B & 1.07s \\
    3D sharding & 23.8B & 1.05s \\
    \bottomrule
  \end{tabular}
\end{table}

\begin{table}
  \caption{Supervised fine-tuning results.}
  \vspace{2mm}
  \label{tab:sft}
  \centering
  \begin{tabular}{c | c | c}
    \toprule
    8.1B/256E win & tie & 12.6B win \\
    \midrule
    44.93\% & 14.87\% & 40.19\%  \\
    \bottomrule
  \end{tabular}
\end{table}

\subsection{Supervised Fine-Tuning Results}
We also study MoE and dense models in the context of instruction-finetuing. Specifically, we finetune the 8.1B/256E MoE and the 12.6B dense pretrain models on an internal dataset to ensure the model is capable of instruction following. Following~\cite{dettmers2024qlora}, we compare the quality of these two models by directly comparing outputs given the same prompt set. To realize this, we  prompt GPT-4 to decide the better response or to announce a tie. Subsequently, the win/loss ratio is calculated and presented in Table~\ref{tab:sft}. The 8.1B/256E MoE shows clear advantage over the 12.6B dense, demonstrating the performance gain on pretrain transfers to SFT.

\subsection{Other Ablations}
\textbf{Router z-loss stabilizes training.} When training large MoEs, e.g. 6.4B/64E, we sometimes hit training instability at the first a few thousand steps. The router z-loss (\cite{zoph2022stmoe}) stabilizes training and does not show any negative impact on model performance and we adopt it in all the model training.

\textbf{A lower expert capacity for training.} We evaluated the modeling trick of using $1.25$ train capacity and $2.0$ eval capacity in ST-MoE (\cite{zoph2022stmoe}) and we found it doesn't help much on the speed-accuracy trade-off curve. For instance, under the 1.6B/256E setting, the trick improves model speed by <5\%, at the expense of hurting CoreEN by 0.2\%. We use a train capacity of $2.0$ in our experiments.

\textbf{Naive second expert routing works.} When selecting the second expert, we found the naive routing that always picks the second best expert works as good as the random routing method used in GShard (\cite{lepikhin2021gshard}). We adopt the naive method in all the experiments.

\begin{table}
\caption{Train compute budget and settings for the main models. Mesh shape is ordered as (\texttt{Data}, \texttt{Expert}, \texttt{Model}).}
\vspace{2mm}
\label{tab:train_budget}
\centering
\begin{tabular}{ l | c  c | c c c } 
\toprule
 model & train tokens & batch size &  TPU devices & mesh shape & step time \\
 \midrule
 6.4B & 128B & 1M & 512 v4 & (1, 256, 1) & 1.69s  \\
 1.6B/256E & 264B & 1M & 512 v4 & (1, 256, 1) & 0.82s \\
 4.8B/256E & 153B & 1M & 512 v4 & (1, 256, 1) & 1.41s \\
 \midrule
 12.6B & 252B & 2M & 1024 v4 & (1, 512, 1) & 2.94s \\
 4.5B/256E & 494B & 2M & 1024 v4 &  (2, 256, 1) & 1.50s \\
 8.1B/256E & 285B & 2M & 1024 v4 & (2, 256, 1) & 2.60s \\
 \midrule
 29.6B & 592B & 2M & 1024 v4 & (1, 512, 1) & 6.56s \\
 6.4B/64E & 2128B & 2M & 1024 v4 & (8, 64, 1) & 1.85s \\
 \bottomrule
\end{tabular}
\end{table}

\section{Conclusion}
In this work, we revisited the step-accuracy trade-off comparisons between MoE and dense LLMs under a challenging setting that favors dense models to MoEs. We first use train step time as a measure for model's computation cost, taking communication overhead in MoE implementations into consideration. Then we adopt the Chinchilla compute-optimal setting, which is optimized for dense model training, as our train compute budget to design comparisons at various scales. We also adopt a 3D sharding method that effectively reduces the communication overhead of MoE implementation. Experimental results show that MoEs consistently outperform dense LLMs under this setting with a meaningful gap on benchmarks including 9 CoreEN 0-shot and 1-shot tasks, MMLU 5-shot and GSM8K 8-shot. 

\newpage
\bibliographystyle{plainnat}
\bibliography{main}

\end{document}